\documentclass[11pt]{article}

\usepackage[utf8]{inputenc}
\usepackage[T1]{fontenc}
\usepackage{times}
\usepackage[margin=1in]{geometry}
\usepackage{hyperref}
\usepackage{url}
\usepackage{graphicx}
\usepackage{booktabs}
\usepackage{caption}
\usepackage{amsmath,amsfonts,amssymb}
\usepackage{bm}
\usepackage{float}

\newcommand{\R}{\mathbb{R}}

\hypersetup{
    colorlinks=true,
    linkcolor=blue,
    citecolor=blue,
    urlcolor=blue
}

\title{Beyond Linearity in Attention Projections: \\ The Case for Nonlinear Queries}

\author{Marko Karbevski \\ In Simplicity Technologies \\ \texttt{marko.karbevski@gmail.com} \\ \texttt{marko.karbevski@insimplicity.tech}}

\date{}

\begin{document}
\maketitle
\let\thefootnote\relax\footnotetext{An earlier version of this paper appeared at the \href{https://gram-workshop.github.io/}{ICLR 2026 Workshop on Geometry-grounded Representation Learning and Generative Modeling (GRaM)}; a substantial part of this work has also been accepted at the \href{https://www.weightsymmetry.com/}{ICML 2026 Workshop on Weight-Space Symmetries}. Code and checkpoints: \url{https://github.com/MarkoKarbevski/beyond_query_linearity}}

\begin{abstract}
Recent algebraic analysis shows that in decoder-only and encoder-only transformers, the Query projection $W_Q$ may be set to identity without noticeable performance deterioration. This is possible because attention depends on $X$ only through the products $XW_Q, XW_K, XW_V$, allowing basis transformations to be absorbed by adjacent layers and propagated through the network. We replace $W_Q \in \R^{d \times d}$ with a nonlinear residual of the form $Q(X) = X + f_\theta(X)$, where $f_\theta$ is a bottleneck MLP with $d^2 + O(d)$ parameters. The identity term anchors the nonlinearity to a known-good prior, supports gradient flow, and preserves expressivity at the chosen bottleneck width. Experiments on GPT-3 small style models show consistent improvement over the baseline ($2.40\%$ lower validation log-loss, $6.81\%$ lower perplexity), comfortably outperforming a model with 12.5\% more non-embedding parameters. This gain is materially larger than what neural scaling laws would predict for the equivalent parameter increase, and the algebraic redundancy of linear $W_Q$ provides a structural reason to expect the modification to remain beneficial at larger scales. The principal purpose of this paper is to motivate testing nonlinear query projections at scale.
\end{abstract}

\section{Introduction}

The transformer architecture \cite{vaswani2017attention} computes attention via Query, Key, and Value projections, each parameterized by learned weight matrices $W_Q, W_K, W_V \in \R^{d \times d}$. Recent work has revealed a fundamental invariance in the attention computation that renders the Query projection algebraically redundant. This observation, grounded in the geometry of the transformer computation graph, motivates a principled modification: replacing the redundant linear projection with a nonlinear one.

\paragraph{Algebraic redundancy.} Consider self-attention transformers with arbitrary masking patterns, which subsumes encoder-only architectures, decoder-only architectures, and certain cross-attention configurations such as prefix language models where the prefix serves as encoder context. Let multi-head attention be defined as $\text{MHA}(X) = \text{Concat}(\text{head}_1, \ldots, \text{head}_h) W_O$, where each head computes $\text{head}_i = \text{Attention}(XW_Q^i, XW_K^i, XW_V^i)$ with $W_Q^i, W_K^i, W_V^i$ the per-head projection matrices. Writing $W_Q = (W_Q^1 | \cdots | W_Q^h)$ and similarly for $W_K, W_V$, attention depends on $X$ only through the products $XW_Q$, $XW_K$, $XW_V$. Karbevski and Mijoski \cite{karbevski2025wkwv} establish that this admits a reparametrization invariance: for any invertible $\Theta$,
\begin{equation}
(X, W_Q, W_K, W_V, W_O) \mapsto (X\Theta, \Theta^{-1}W_Q, \Theta^{-1}W_K, \Theta^{-1}W_V, W_O)
\end{equation}
leaves the MHA output unchanged. Since singular matrices have Lebesgue measure zero, $W_Q$, $W_K$, and $W_V$ are almost surely invertible, allowing the choice $\Theta = W_Q$ (or $W_K$, $W_V$) to simplify one of the three matrices. Under mild conditions on the network topology, this basis change propagates through the entire network: each layer passes its transformation to its predecessor, telescoping back to the embedding layer. Graef \cite{graef2024transformer} first proved this propagation for simplified architectures; Karbevski and Mijoski \cite{karbevski2025wkwv} strengthened it along complementary axes (single-layer, attention-skip-only, and weight-shared multi-layer settings), investigated in detail the pitfalls of extension through layer normalization, and empirically verified $W_Q = I$ in standard transformers. Setting $\Theta = W_Q$ yields $W_Q \mapsto I$, establishing that $W_Q$ is algebraically redundant. Crucially, this extends to Grouped Query Attention: since GQA shares $W_K$ and $W_V$ across query groups, only $W_Q$ can be eliminated without disrupting the shared structure, making queries the natural choice. The invariance is also compatible with Mixture-of-Experts. While we test with learned positional embeddings, this reparametrization also holds for RoPE \cite{su2024roformer}, as shown by Karbevski and Mijoski \cite{karbevski2025wkwv}.

Empirically, models trained with $W_Q = I$ match baseline performance \cite{karbevski2025wkwv}, confirming the identity as a good prior for queries, both for representability and for training stability. Combined with the algebraic redundancy, this leaves nonlinearity as the only meaningful way to allocate parameters to the query pathway: any linear $W_Q$ is absorbed into adjacent layers and changes nothing.

\paragraph{Why queries, not keys or values.} The projections $W_Q$, $W_K$, $W_V$ are not interchangeable. Within each head, if $W_Q^i = W_K^i$, then $Q^i {K^i}^T = XW_Q^i {W_Q^i}^T X^T$ is symmetric; empirically, trained attention matrices are not symmetric. Hu et al.\ \cite{hu2021lora} report that low-rank adaptation of $W_Q$ and $W_V$ jointly yields superior fine-tuning performance, whereas adapting $W_K$ alone is insufficient. Li et al.\ \cite{li2023topic} observe two-stage training: $\|W_V\|_F$ grows while $\|W_Q\|_F, \|W_K\|_F \approx 0$ in early training; only later do $W_Q, W_K$ begin learning. These findings motivate giving the query pathway its own dedicated computation.

One might ask whether the same residual structure could apply to K and V. Two considerations make this a more involved modification. \emph{Structurally}, symmetric skip structure in both Q and K introduces a parameter-free $XX^T$ term biasing attention toward self-similarity. Moreover, Ji et al.\ \cite{ji2025skipattention} show that self-attention is uniquely ill-conditioned and dependent on skip connections for regularization; removing the skip around attention, as explored by He et al.\ \cite{he2023vanillatransformers}, might help but requires careful initialization. Asymmetric designs using Hyper-Connections \cite{zhu2024hyperconnections, xie2025mhc} may resolve this tension. \emph{From a parameter-budget perspective}, $W_Q$ is uniquely absorbable, which is what allows us to replace it at no net parameter cost. $W_K$ and $W_V$ are not absorbable in the same way, so a naive nonlinearization would have to retain $W_K$ and $W_V$ as trainable matrices and \emph{add} nonlinear components on top, materially increasing the parameter count. This may well be worth the additional budget, but the comparison would no longer be parameter-neutral and would warrant a different experimental design. The structural and parameter-budget questions for K and V, together with systematic ablations on the shape and parameter budget of $f_\theta$ (bottleneck width, activation choice, and the placement of normalization layers, in particular whether $f_\theta$'s normalization should be coupled with the $W_Q$ replacement as in the current design or operate as a separate stage), constitute important future work.

\paragraph{Proposal.} Following the suggestion of Karbevski and Mijoski \cite{karbevski2025wkwv}, we implement a nonlinear query projection directly:
\begin{equation}\label{eq:proposed}
Q(X) = (X + f_\theta(X))/2
\end{equation}
where $f_\theta$ is a bottleneck MLP with $d^2 + O(d)$ learnable parameters, the same order as the $W_Q$ it replaces. The $1/2$ scaling follows Karbevski and Mijoski \cite{karbevski2025wkwv}. The identity term $X$ anchors the nonlinearity to a known-good prior, provides a direct path for gradient flow \cite{he2016deep}, and is necessary to recover full-rank expressivity from the bottleneck: since $f_\theta$ factors through a $d/2$-dimensional intermediate representation, $\dim_H(\mathrm{Im}\, f_\theta) \le d/2$ in Hausdorff dimension, while $\sup_\theta \dim_H(\mathrm{Im}(\mathrm{Id} + f_\theta)) = d$, with the image attaining all of $\mathbb{R}^d$. Keys and values remain standard linear projections. Specifically, we implement $f_\theta$ as:
\begin{equation}
f_\theta(X) = \text{LN}\left(\text{GELU}\left(\text{RMSNorm}(X) W_1\right) W_2\right)
\end{equation}
where $W_1 \in \R^{d \times r}$, $W_2 \in \R^{r \times d}$, and $r = d/2$. The query $Q(X) = (X + f_\theta(X))/2$ is applied independently per token. The attention logits $Q^i {K^i}^T / \sqrt{d_k}$ per head use standard scaling. Matrix parameters total $2dr = d^2$; normalization layers add $O(d)$ parameters (${\sim}0.03\%$ overhead). The design prioritizes stability over throughput or final quality; we conjecture that this formulation may be overly conservative. The choice of LN and RMSNorm was made for stability in initial experiments; systematic exploration of normalization variants, or their elimination entirely, is an immediate priority for future work.

\paragraph{Related work.} Kernel-based attention methods \cite{choromanski2020rethinking, katharopoulos2020linear} apply nonlinear feature maps $\phi(Q), \psi(K)$ to achieve linear complexity, but retain standard linear projections $Q = XW_Q$ before the nonlinearity. MLP-Attention \cite{mlpattention} eliminates Q and K projections entirely, replacing the $QK^T$ computation with an MLP that maps embeddings directly to attention weights; however, this adds substantial parameters (${\sim}10\%$), was validated only at small scale (${\sim}5$M parameters on character-level Tiny Shakespeare), uses no residual structure, and lacks theoretical motivation. Zhang \cite{zhang2023neural} replaces all of Q, K, V with two-layer feedforward MLPs (no residual), also adding substantial parameters; they fine-tune pretrained models rather than training from scratch, and offer no algebraic justification. Nonlinear extensions of LoRA \cite{li2024loran, ji2024sinelora, dong2025aurora} introduce nonlinearities to break the low-rank bottleneck of adapters in parameter-efficient fine-tuning; this is a distinct setting from base architecture design, with different goals and justification. Concurrent work by Qiu et al.\ \cite{qiu2025gatedattention} introduces nonlinearity on the output side of attention via a sigmoid gate applied after the $QK^T V$ computation; their motivation and placement differ from ours, but the two modifications are complementary and could in principle be combined. Building on the suggestion of Karbevski and Mijoski \cite{karbevski2025wkwv}, we present the first direct implementation of nonlinear query projections: a principled, parameter-neutral architectural modification with residual structure for pretraining.

\section{Experiments}

\subsection{Setup}

We train on OpenWebText on a single NVIDIA RTX 5090 GPU. The baseline is NanoGPT \cite{nanogpt}: 12 layers, 12 heads, $d_{\text{model}} = 768$, MLP hidden dimension $3072 = 4d_{\text{model}}$, context length 1024, no biases, tied embedding/LM-head, GPT-2 tokenizer. We write $d = d_{\text{model}}$ throughout. We use AdamW with $\beta_1 = 0.9$, $\beta_2 = 0.95$, 2k steps of warmup, cosine decay, and gradient clipping at 1.0; ${\sim}$490k tokens per gradient step. Data splits use NanoGPT's provided seeds. To ensure fair comparison, we pre-generate all batch indices from a fixed random seed; every model sees identical training data at each step and identical evaluation data at each validation step. Validation loss is estimated every 1000 steps by averaging over 2400 sequences (${\approx}$2.5M tokens).

Training runs for 60k steps (${\sim}$29B tokens), far beyond Chinchilla-optimal \cite{hoffmann2022chinchilla} for 124M parameters (${\sim}$2.5B tokens). This over-training regime has been shown to scale reliably and predictably with respect to extrapolations from smaller-scale fits \cite{gadre2024scaling}, and we use it here to test whether improvements persist under extended training, representing ``hard'' wins rather than gains that might vanish with proper token scaling.

We compare \textbf{Residual GELU} ($Q = (X + f_\theta(X))/2$) against the \textbf{Baseline} and MLP-widened controls (MLP hidden dimension $4.75d$ instead of $4d$, adding 12.5\% non-embedding parameters). This direct comparison isolates architectural benefits from capacity gains, avoiding reliance on scaling law extrapolations \cite{kaplan2020scaling, hoffmann2022chinchilla}. The baseline uses weight decay $0.1$ and learning rate $6 \times 10^{-4}$ decaying to $6 \times 10^{-5}$. Given the stability observed with $W_Q = I$ \cite{karbevski2025wkwv}, we explored higher peak learning rates (up to $4 \times 10^{-3}$) and weight decay values from the standard $0.10$ down to $2^{-5} \approx 0.03$ for the nonlinear variant. Headline configurations are reported in Table~\ref{tab:results}; the broader sweep (including $\lambda = 2^{-5}$) appears in Figure~\ref{fig:sweep} of the appendix.

\subsection{Results}

\begin{table}[H]
\centering
\small
\caption{Configurations and validation loss at step 60k. MLP$_w$ denotes the baseline with MLP hidden dimension $w{\cdot}d$. The $\Delta$ N.E.\ Params column (N.E.\ = non-embedding) reports the change in parameter count relative to baseline; the bottleneck MLP $f_\theta$ is sized to match $W_Q$ at $d^2$ parameters, leaving only a small LayerNorm overhead. The $\Delta$\,LL and $\Delta$\,PPX columns report relative improvement over baseline: $\Delta\text{LL} = (\mathcal{L}_{\text{base}} - \mathcal{L}) / \mathcal{L}_{\text{base}}$ and $\Delta\text{PPX} = 1 - \exp(\mathcal{L} - \mathcal{L}_{\text{base}})$.}
\label{tab:results}
\vspace{0.5em}
\begin{tabular}{lccccccc}
\toprule
Config & $\Delta$ N.E.\ Params & $\lambda$ (WD) & $\eta_{\max}$ & $\eta_{\min}$ & Val loss & $\Delta$\,LL & $\Delta$\,PPX \\
\midrule
Baseline                 & ref.     & $0.10$    & $6{\cdot}10^{-4}$      & $6{\cdot}10^{-5}$    & $2.9441$ & ref.           & ref.           \\
MLP$_{4.75}$             & $+12.5\%$ & $0.10$   & $6{\cdot}10^{-4}$      & $6{\cdot}10^{-5}$    & $2.9165$ & $+0.94\%$      & $+2.72\%$      \\
MLP$_{4.75}$ (scaled)    & $+12.5\%$ & $0.10$   & $5.66{\cdot}10^{-4}$   & $5.66{\cdot}10^{-5}$ & $2.9173$ & $+0.91\%$      & $+2.65\%$      \\
Res.\ GELU               & ${\sim}0.03\%$ & $0.05$   & $2.4{\cdot}10^{-3}$    & $3{\cdot}10^{-5}$    & $2.9015$ & $+1.45\%$      & $+4.17\%$      \\
Res.\ GELU               & ${\sim}0.03\%$ & $0.10$   & $4.0{\cdot}10^{-3}$    & $1{\cdot}10^{-5}$    & $2.8750$ & $+2.35\%$      & $+6.68\%$      \\
Res.\ GELU (best)        & ${\sim}0.03\%$ & $0.10$   & $3.6{\cdot}10^{-3}$    & $2{\cdot}10^{-5}$    & $\mathbf{2.8736}$ & $\mathbf{+2.40\%}$ & $\mathbf{+6.81\%}$ \\
\bottomrule
\end{tabular}
\end{table}

\begin{figure}[H]
\centering
\includegraphics[width=0.65\textwidth]{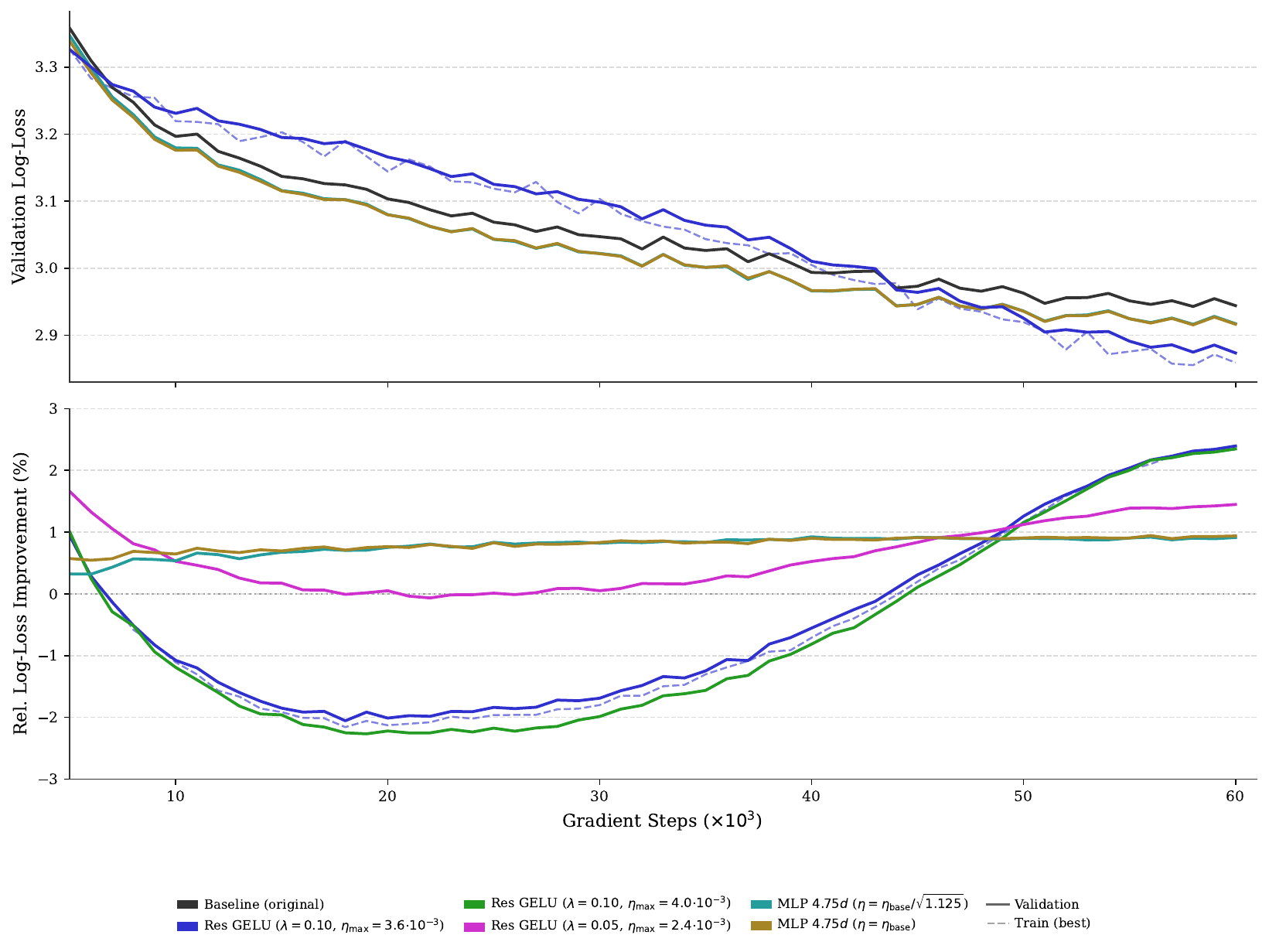}
\caption{Training dynamics (steps 5k--60k). Solid: validation; dashed: training curve for the best configuration. \emph{Top:} absolute validation log-loss for baseline, both MLP$_{4.75}$ variants, and the best Res.\ GELU. \emph{Bottom:} relative log-loss improvement over baseline for all five non-baseline configurations.}
\label{fig:all}
\end{figure}

Figure~\ref{fig:all} and Table~\ref{tab:results} summarize the results. The best nonlinear query ($\lambda=0.10$, $\eta_{\max}=3.6{\cdot}10^{-3}$) reaches $2.8736$ vs $2.9441$ for baseline at 60k steps, a $2.40\%$ log-loss ($6.81\%$ perplexity) improvement, comfortably outperforming MLP$_{4.75}$ ($+0.94\%$) which has 12.5\% more non-embedding parameters. The two high-LR Res.\ GELU variants reach near-identical final loss. For reference, the rotary-vs-learned-absolute improvement reported by EleutherAI \cite{biderman2021rope} was $+1.78\%$ log-loss / $+4.88\%$ perplexity ($2.809 \to 2.759$), measured on a 125M-parameter GPT-NeoX model \cite{black2022gptneox} with GPT-3 small hyperparameters, trained for 55k steps (${\sim}$30B tokens) on OpenWebText2 \cite{gao2021pile}, a deduplicated and quality-filtered corpus cleaner than the OpenWebText used here. Relative improvement is non-monotone (Figure~\ref{fig:all}, bottom), though absolute loss descends monotonically for all configurations (top panel).

The best configuration uses \emph{standard} weight decay ($\lambda=0.10$) with a peak LR of $3.6{\cdot}10^{-3}$ ($6\times$ the baseline). The architecture also trains stably at reduced weight decay down to $2^{-5} \approx 0.03$ (see Figure~\ref{fig:sweep} in the appendix), whereas the baseline diverges before 20k steps at $\lambda=0.05$; however, lower WD requires a correspondingly lower peak LR: $\lambda=0.05$ is stable only up to $\eta_{\max}=2.4{\cdot}10^{-3}$ and yields only $+1.45\%$. This complements findings of Andriushchenko et al.\ \cite{andriushchenko2024wd} on weight decay necessity, whose NanoGPT experiments tested only $\lambda\in\{0.1, 0\}$; we fill the gap in between. The tolerance of higher learning rates may relate to scale-invariance from normalization layers \cite{lyle2024normalization}.

\section{Discussion}

\paragraph{Limitations.} Our experiments use a single model scale (${\sim}$124M parameters) with 60k training steps and no systematic hyperparameter search. The hyperparameter landscape (learning rate schedules, weight decay, bottleneck architecture, LayerNorm epsilon) remains largely unexplored, suggesting that the reported $2.40\%$ improvement represents a floor rather than a ceiling. Inference speed has not been measured. Multiple seeds have not been run; we mitigate this by training and evaluating all models on identical batches, and by training far beyond Chinchilla-optimal \cite{hoffmann2022chinchilla} (${\sim}$29B tokens versus ${\sim}$2.5B optimal), a regime that has been shown to scale reliably and predictably with respect to extrapolations from smaller-scale fits \cite{gadre2024scaling}, reducing the influence of early stochasticity. Larger scales and weight dynamics analysis remain for future work; we do not report explicit downstream evaluation, though perplexity is established as a strong predictor of downstream performance at this scale: \cite{gadre2024scaling} fit a power law from perplexity to aggregate top-1 error across a downstream task suite, and \cite{xia2023training} show that perplexity predicts in-context-learning performance on 74 BIG-Bench multiple-choice tasks independent of model size, including the $125$M-parameter regime of our experiments.

\paragraph{Efficiency and deployment.} The bottleneck MLP replaces $W_Q$ at the same parameter count and FLOPs; overhead from the GELU nonlinearity and normalization layers is negligible. However, the nonlinear query introduces a serial dependency: the bottleneck MLP must complete before attention can proceed. Custom kernel implementations may be needed for practical deployment. Further efficiency is possible: since the first bottleneck layer maps $d \to r = d/2$, it can be fused with the Key and Value projections, reducing the combined $W_{QKV}$ output from $d \to 3d$ to $d \to 2.5d$. ReLU or ReLU$^2$ activations should be tested as alternatives to GELU for additional speed gains.

\paragraph{Scaling.} As $d$ increases, the bottleneck dimension $r = d/2$ grows proportionally, increasing the expressivity of $f_\theta$. The observed gain at 124M ($2.40\%$ log-loss) is materially larger than the marginal improvement that scaling laws \cite{kaplan2020scaling, hoffmann2022chinchilla} would predict for the equivalent parameter increase: in our own control, $12.5\%$ more non-embedding parameters yields only $0.94\%$, and Chinchilla-style fits at this regime extrapolate similar marginal returns. Inverting this rate: matching the nonlinear query's $2.40\%$ log-loss gain with the baseline architecture would require ${\sim}35\%$ more non-embedding parameters at $124$M scale (fitting a local Chinchilla exponent to the $0.94\%/{+}12.5\%$ data point), substantially exceeding the seed-level stochasticity expected in this $12{\times}$-Chinchilla-optimal overtrained regime, where loss trajectories have been shown to scale reliably and predictably \cite{gadre2024scaling}. Whether the gap reflects a genuine architectural advantage that persists at scale, or shrinks under standard scaling-law decay, is an empirical question that motivates this paper's central call: testing the modification at larger scales. Unlike Primer \cite{so2021primer}, which achieved improvements through depthwise convolutions with explicit locality bias (a bias larger models learn naturally, making the improvement redundant at scale), there is no obvious path by which scaling would make our approach redundant: the identity anchor provides a natural skip connection, the bottleneck grows with model size, and the algebraic redundancy of linear $W_Q$ is itself scale-independent.

\paragraph{Extensions.} Our results suggest that Karbevski and Mijoski \cite{karbevski2025wkwv} may benefit from further learning rate increases: K\&M report that the reduced architecture tolerates $2.7\text{--}3.7\times$ the baseline learning rate, attributing this to attention logits becoming linear rather than quadratic in the learned weights; we find the nonlinear variant extends this further, operating stably at $5\times$ baseline. Systematic comparison with output-side gating mechanisms \cite{qiu2025gatedattention} is another direction. The identity anchor makes our approach compatible with RoPE, MoE, and GQA/MQA \cite{karbevski2025wkwv, graef2024transformer}. Rather than a separate $f_\theta$, one could derive queries from intermediate MLP activations ($f_\theta(X) = \text{MLP}_{\text{hidden}}(X)_{[0:d]} W'$), reusing computation; a structured generalization, widening the MLP output with per-pathway and pair-shared concept subspaces, is sketched in Appendix~\ref{sec:geometric}. Extending the residual nonlinear construction to $W_O$ is the natural symmetric move; Graef \cite{graef2024transformer} establishes the algebraic redundancy of $W_O$ in the simultaneously skipless and LayerNorm-free regime, where this extension applies. For pretrained models, one could anchor to existing projections: $Q = (XW_Q + f_\theta(X))/2$; this relates to nonlinear LoRA extensions \cite{li2024loran, ji2024sinelora, dong2025aurora}.

\paragraph{Toward a geometric and mechanistic theory.} The empirical gap between the observed gain and scaling-law predictions (Scaling paragraph above) motivates a theoretical explanation that the algebraic redundancy alone does not provide. Appendix~\ref{sec:geometric} sketches two preliminary attempts: an informal concept-space picture rooted in the Linear Representation Hypothesis (with two roadblocks linking pathway-agnostic MLPs and the absence of downstream nonlinear combination), and a formal affine-hull bound on the QKV image (any linear pathway constrains the image to a flat $d$-dim subspace of $\mathbb{R}^{3d}$, leaving $2d$ ambient dimensions unreachable). The first is mechanistic-interpretability-flavored and would benefit from direct probing of trained $f_\theta$ to test whether feature-decoupling actually occurs; the second is coordinate-free and assumption-light but currently states an upper bound rather than a sharp characterization. Neither explanation predicts the magnitude of the observed effect; together they identify structural reasons the effect should be nonzero. A natural follow-up direction is a quantitative theory linking the curvature of $f_\theta$'s image to its affine hull dimension and, ultimately, to representational capacity in the attention pathway.

\section{Conclusion}

We presented the first direct implementation of nonlinear query projections as suggested by Karbevski and Mijoski \cite{karbevski2025wkwv}: a residual form $Q(X) = (X + f_\theta(X))/2$ that anchors nonlinearity to the identity prior. This design is motivated by the observation that linear $W_Q$ is redundant under mild conditions, yet queries may benefit from dedicated nonlinear computation decoupled from the value stream. At the same parameter budget, the nonlinear query achieves $2.40\%$ lower validation log-loss ($6.81\%$ lower perplexity) than baseline, comfortably outperforming a model with 12.5\% more non-embedding parameters. The modification also improves training stability, tolerating weight decay and learning rate settings where the baseline diverges. The gain at 124M materially exceeds what scaling laws would predict for the equivalent parameter increase, and the algebraic redundancy argument provides a structural reason to expect the effect to persist at larger scales. Testing the modification at scales beyond our 124M setting is the priority next experiment; our formulation prioritizes stability over throughput or final quality, suggesting these results represent a floor rather than a ceiling.

\section*{Acknowledgements}

I am grateful to the anonymous reviewers for their constructive feedback, which improved the clarity of this work. I also thank Nils Graef, Yiping Ji, Haris Mandal, and Antonij Mijoski for valuable discussions.

\appendix
\section{Hyperparameter Sweep}
\label{sec:sweep}

\begin{figure}[H]
\centering
\includegraphics[width=0.92\textwidth]{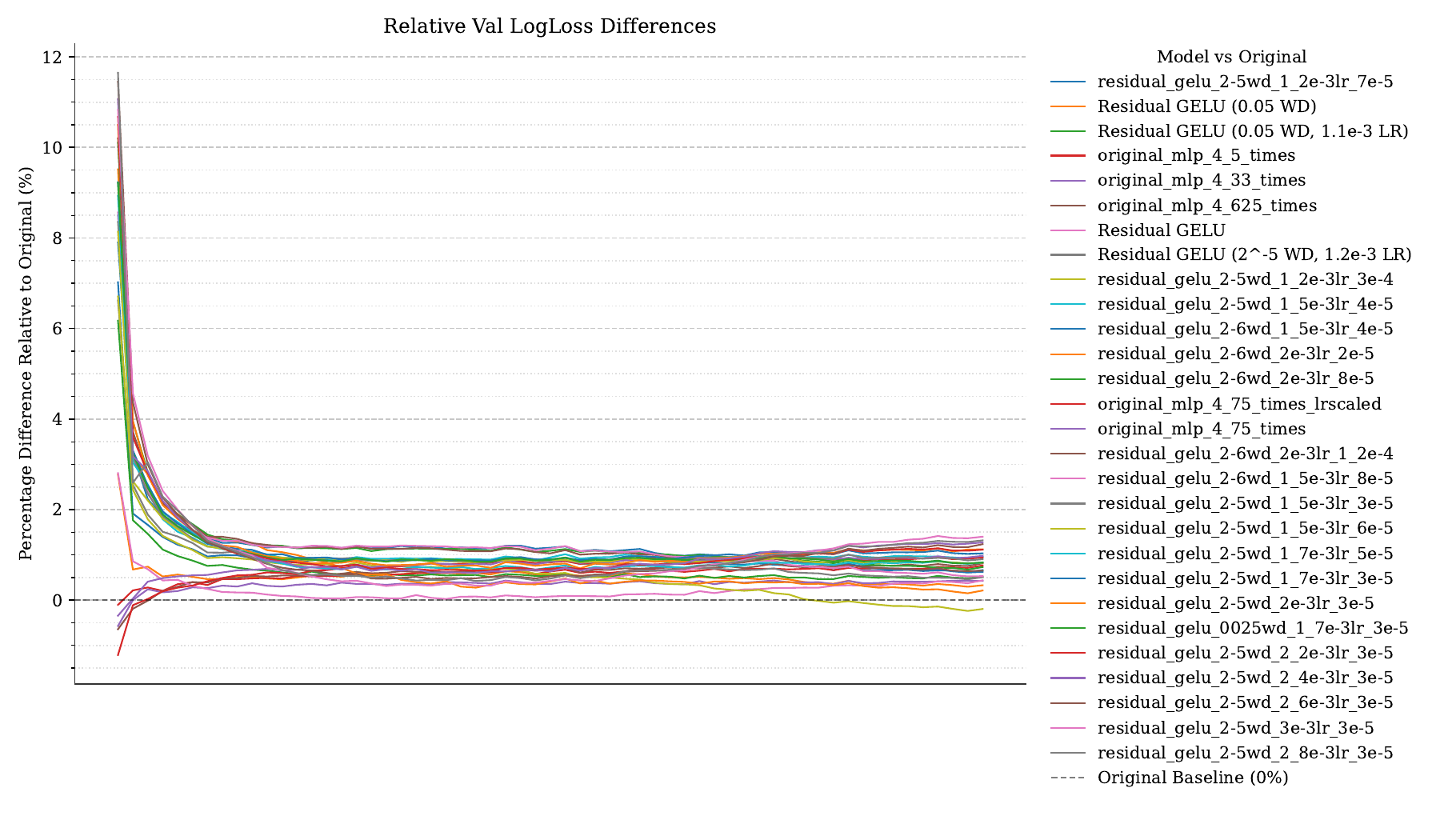}
\caption{Relative improvement over baseline (steps 1k to 59k). Nonlinear configurations: 84.97M parameters; MLP-widened controls: 89.6 to 95.6M.}
\label{fig:sweep}
\end{figure}

This section is illustrative and does not claim completeness. The figure below is an earlier snapshot of the sweep, covering roughly 70\% of the hyperparameter configurations we ran; the complete set of runs and checkpoints is available in the code repository. We include it to expose the sweep conducted and to highlight the unexpectedly high initial gains across configurations, which stabilize mid-training (Figure~\ref{fig:sweep}).

\section{On the Limitations of the Linear $\mathbb{R}^d \to \mathbb{R}^{3d}$ QKV Pathway}
\label{sec:geometric}

The empirical gain from replacing linear $W_Q$ with the nonlinear residual $f_\theta$ ($2.40\%$ lower validation log-loss at 124M parameters) is materially larger than what neural scaling laws \cite{kaplan2020scaling, hoffmann2022chinchilla} predict for the equivalent parameter increase (our own control measures $0.94\%$ for a $12.5\%$ MLP widening; Chinchilla-style fits at this regime extrapolate similar marginal returns), and exceeds what reasonable seed-level stochasticity could account for. The algebraic redundancy of Karbevski and Mijoski \cite{karbevski2025wkwv} and Graef \cite{graef2024transformer} explains why linear $W_Q$ is \emph{wasted}, but does not by itself explain why a parameter-equivalent nonlinear replacement should outperform a similarly-sized linear architecture by such a margin.

This appendix sketches two complementary, preliminary perspectives that attempt to formalize this gap. The first is informal and rests on a Linear-Representation-Hypothesis-style assumption \cite{park2023linear, elhage2022superposition}; the second is a formal observation about the affine geometry of the QKV map and requires no such assumption. Both are speculative and deserve more careful development than we give them here; in particular, neither is sufficient on its own to predict the magnitude of the observed gain, only to identify structural reasons it should be nonzero. We include them to make the conjecture concrete and to provide a starting point for follow-up investigation.

\subsection{The Concept-Space Picture (informal)}
\label{sec:geometric-informal}

Suppose, in the spirit of the Linear Representation Hypothesis, that the immediate upstream MLP+Id block writes features into the residual stream along learned concept directions $(c_i)_{i \in I}$. The input to the attention's QKV projection can then be written as $y = \sum_{i \in I} a_i\, c_i$, with concept coefficients $(a_i)$ that depend on the token. Under this view, the linear $W_{QKV} = [W_Q; W_K; W_V]$ pathway faces two related structural difficulties.

\paragraph{Roadblock 1: One $y$ must serve three pathways.} By linearity, $W_P\, y = \sum_i a_i\, (W_P c_i)$ for each $P \in \{Q, K, V\}$. The token-dependent coefficients $a_i$ are identical across the three pathways; each $W_P$ can only re-express the concept basis as $(W_P c_i)$ but cannot reweight $a_i$ pathway-by-pathway. The upstream MLP+Id is therefore solving an encoding problem: pack into $d$ dimensions a concept activation pattern that simultaneously serves three distinct downstream uses (querying, being queried, contributing value).

\paragraph{Roadblock 2: No nonlinear combination downstream.} Any nonlinear combination of concept activations that any of the three pathways might need must be precomputed by the MLP+Id and encoded directly in $y$, because the path from $y$ through $W_{QKV}$ to the attention softmax is purely linear. If the query pathway needs to attend conditionally on a conjunction of concepts (e.g., ``anger and question''), that conjunction must be allocated its own concept dimension in $y$ at the MLP+Id stage. There is no architectural component between the MLP+Id and the attention softmax that can synthesize such conjunctions from constituent activations.

Together, the two roadblocks suggest that pathway-specific nonlinearity placed between the MLP+Id and attention addresses both at once: it allows pathway-specific concept reweighting (Roadblock~1) and pathway-specific nonlinear combination (Roadblock~2). The nonlinear residual query proposed in this work is one realization. A natural alternative reuses the upstream MLP+Id's already-computed nonlinear hidden state $\mathrm{GELU}(W_{\mathrm{up}}\, \cdot) \in \mathbb{R}^{4d}$: it contains $4d$ nonlinearly-computed features of its input, of which only a single $d$-dimensional linear combination reaches the next attention via $W_{\mathrm{down}}$. Routing this hidden state (or a $d$-dim slice of it) directly into the next attention's query pathway, concretely $Q(X) = (X + h^{(\ell-1)} W'_Q)/2$ where $h^{(\ell-1)} = \mathrm{GELU}(W_{\mathrm{up}}^{(\ell-1)}\, \cdot)$ is the previous block's MLP hidden state, gives the query access to a $4d$-dim nonlinear basis that the $K$ and $V$ pathways do not see (addressing Roadblock~1) and that is already nonlinearly composed (addressing Roadblock~2), at the cost of one additional linear projection and no recomputed nonlinearity. The proposed $f_\theta$ is the more conservative, intra-block realization of the same architectural intuition; the MLP-reuse variant is left to future work.

A more aggressive refinement of the MLP-reuse design widens the MLP output to
\[
\mathbb{R}^{d + d_Q + d_K + d_V + d_{Q,K} + d_{Q,V} + d_{K,V}},
\]
where the standard $d$-dim component continues to feed the residual stream and the remaining subspaces are routed to specific pathways of the next attention block: per-pathway dimensions $d_Q, d_K, d_V$ are visible only to a single pathway, and pair-shared dimensions $d_{Q,K}, d_{Q,V}, d_{K,V}$ are visible to a pathway pair (so the attention-logit pathway sees $d + d_{Q,K}$ shared dimensions plus its individual $d_Q$ or $d_K$, and so on). This is the most explicit architectural realization of pathway-specific concept access from Roadblock~1: each pathway is allocated its own designated concept subspace by construction, with no shared $W_Q, W_K, W_V$ projection forcing the same basis on all three. The allocation of dimensions across the seven subspaces is a design choice that requires empirical investigation, as does the integration with attention-head structure; we leave both the architecture and the training dynamics to future work.

This picture rests on a substantive assumption: that something like the Linear Representation Hypothesis holds at the relevant layer and architectural depth, and that the MLP+Id is the operative concept-writing component. Both claims have empirical support from the mechanistic interpretability literature but neither is established for the specific architecture studied here. Whether trained $f_\theta$ in fact performs pathway-specific concept reweighting (as opposed to other forms of useful computation) is an empirical question we leave open. Further development of this picture, including mechanistic probing of trained $f_\theta$ and quantitative tests of the roadblocks, is a natural direction for follow-up work.

\paragraph{Geometric restatement.} The two roadblocks above admit a coordinate-free restatement. The upstream MLP+Id, however expressive its nonlinear computation, can only produce content that the downstream linear $W_{QKV}$ pathway then projects into a $d$-dimensional linear subspace of the $3d$-dimensional ambient space inhabited by $(Q, K, V)$. The MLP+Id is therefore bottlenecked not only by its own output width $d$ but, downstream, by the fact that the linear $W_{QKV}$ map cannot reach more than a $d$-dim slice of the available $3d$ ambient capacity; the other $2d$ ambient directions are unreachable by \emph{any} choice of linear projections.

This restriction is materially stronger than ``$d$-dimensional image in a $3d$-dimensional ambient'' might suggest. A $d$-dim linear subspace is flat: every output point lies in the same fixed $d$-dim slice, regardless of input. A $d$-dim curved submanifold (or even a topologically $d$-dim set produced by a nonlinear map) can wind through $\mathbb{R}^{3d}$ with its affine hull occupying up to the full $3d$ ambient capacity; a $1$-dim helix in $\mathbb{R}^3$ is the canonical illustration, locally $1$-dimensional yet globally spanning all three ambient axes. In representational terms, two $d$-dim sets with the same topological dimension can differ substantially in which regions of $\mathbb{R}^{3d}$ they can reach. The linear $W_{QKV}$ image is the most restrictive case. The next subsection makes this geometric constraint precise without invoking the LRH-style assumption used above.

\subsection{The Affine Hull Bound (formal)}
\label{sec:geometric-formal}

A complementary observation concerns the geometric extent of the QKV pathway's image in $\mathbb{R}^{3d}$, without reference to learned features or concept bases.

\paragraph{Definition (affine hull dimension).} For a set $S \subseteq \mathbb{R}^N$, the affine hull dimension is
\[ \dim\, \mathrm{aff}(S) := \dim\,\mathrm{span}\{s - s_0 : s \in S\} \qquad (\text{for any fixed } s_0 \in S), \]
i.e., the dimension of the smallest affine subspace containing $S$. This is distinct from the topological dimension of $S$ as a manifold and from its Hausdorff dimension. A $1$-dim helix in $\mathbb{R}^3$ has topological and Hausdorff dimension $1$ but affine hull dimension $3$: locally it has one degree of freedom, but globally it spans the full ambient space. A space-filling curve (Peano, Hilbert) shows the extreme: $1$-dimensional continuous parametrization, affine hull equal to the entire ambient $\mathbb{R}^k$.

\paragraph{Setup.} Consider a generalized QKV-style map
\[ \Phi: \mathbb{R}^d \to \mathbb{R}^{3d}, \qquad \Phi(x) = (f(x),\, Px,\, Qx), \]
where $f: \mathbb{R}^d \to \mathbb{R}^d$ is possibly nonlinear and $P, Q: \mathbb{R}^d \to \mathbb{R}^d$ are linear. The standard fully linear QKV pathway corresponds to $f$ also linear; the architecture proposed in this work corresponds to $f$ nonlinear with $P, Q$ retained as the linear key and value projections.

\paragraph{Claim.} Let $S = \Phi(\mathbb{R}^d)$. Then
\[ \dim\,\mathrm{aff}(S) \;\le\; \dim\,\mathrm{span}\bigl(\mathrm{im}(f) - f(0)\bigr) \;+\; \mathrm{rank}([P; Q]) \;\le\; 2d, \]
where $[P; Q] \in \mathbb{R}^{2d \times d}$ is the vertical stacking of $P$ and $Q$.

\paragraph{Proof.} Let $W = \mathrm{span}\{\Phi(x) - \Phi(0) : x \in \mathbb{R}^d\}$, so $\dim\,\mathrm{aff}(S) = \dim W$. Decompose $\mathbb{R}^{3d} = \mathbb{R}^d \oplus \mathbb{R}^{2d}$ into the $f$-coordinates and the $(P, Q)$-coordinates, with projections $\pi_1, \pi_2$. From $\Phi(x) - \Phi(0) = (f(x) - f(0),\, Px,\, Qx)$,
\[
\pi_1(W) \;\subseteq\; \mathrm{span}\bigl(\mathrm{im}(f) - f(0)\bigr), \qquad \pi_2(W) \;=\; \mathrm{im}([P; Q]).
\]
Applying the sub-additivity bound $\dim W \le \dim \pi_1(W) + \dim \pi_2(W)$ (rank-nullity for $\pi_2|_W$, with $\ker(\pi_2|_W) \subseteq \pi_1(W) \oplus 0$) gives the claim. $\hfill\square$

The bound treats $\pi_1$ and $\pi_2$ as independent, while both depend on the same $x$, and is therefore loose when this coupling is strong. The extreme case is linear $f$: then $\Phi$ is itself a single linear map $\mathbb{R}^d \to \mathbb{R}^{3d}$ of rank at most $d$, so $\dim\,\mathrm{aff}(S) \le d$, half of what the sub-additivity bound asserts. The bound is approached only when $f$ is nonlinear in a way that lets $\pi_1(W)$ contribute directions decoupled from $\pi_2(W)$; a tight characterization of this coupling is left to future work.

\paragraph{Three architectural regimes.} Specializing to the three placements of nonlinearity in the QKV pathway:
\begin{itemize}
\item Fully linear pathway ($f$ also linear): the entire map $\Phi$ is linear and $\dim\,\mathrm{aff}(S) \le d$. The image is contained in a $d$-dim linear subspace of $\mathbb{R}^{3d}$, leaving $2d$ ambient dimensions unreached.
\item One nonlinear pathway ($f$ nonlinear, $P$ and $Q$ linear, the architecture in this work): $\dim\,\mathrm{aff}(S) \le 2d$, with equality approached for sufficiently nonlinear $f$.
\item All three nonlinear: $\dim\,\mathrm{aff}(S) \le 3d$ by analogous argument, with similar conditions for tightness.
\end{itemize}
The qualitative point is that moving from the all-linear case to the one-nonlinear case can unlock up to $d$ additional ambient dimensions of geometric extent, conditional on $f$ being substantively nonlinear. The marginal return on additional nonlinear pathways is similarly conditional and not automatic.

\paragraph{Interpretation and limitations.} The upper bound rests on linear-algebraic and finite-difference facts and requires no assumption about learned features or the structure of the residual stream. However, it is an upper bound, not an achievability statement, and a precise characterization of when nonlinearity actually unlocks additional ambient dimensions (in particular, a clean quantitative measure of ``substantive nonlinearity'' relating to the variation of $f$ across the input distribution) is left to future work. Whether trained $f_\theta$ actually exploits this geometric headroom, and whether the geometric gain corresponds to the empirical performance improvement we observe, are open empirical questions.

\end{document}